\documentclass[jou, a4paper, hidelinks]{apa7}

\usepackage{CJKutf8}

\usepackage[american]{babel}
\usepackage{float} 
\usepackage[T1]{fontenc}
\usepackage[utf8]{inputenc} 
\usepackage{inconsolata}
\usepackage{fvextra} 
\DefineVerbatimEnvironment{prompt}{Verbatim}{
  breaklines=true,  
  breakanywhere=true,  
  breaksymbolleft={},    
  fontsize=\small,   
  obeytabs=true,
  tabsize=2
}

\usepackage{csquotes}
\usepackage[style=apa,sortcites=true,sorting=nyt,backend=biber]{biblatex}
\DeclareLanguageMapping{american}{american-apa}
\title{The ``Small World of Words'' German Free-Association Norms}
\shorttitle{SWOW-DE}

\authorsnames[{1, 2}, 1, 3, 4, {2, 1}]{Samuel Aeschbach, Rui Mata, Kaidi Lõo, Simon De Deyne, Dirk U. Wulff}
\authorsaffiliations{
    {Center for Cognitive and Decision Sciences, Faculty of Psychology, University of Basel, Switzerland},
    {Center for Adaptive Rationality, Max Planck Institute for Human Development, Berlin, Germany},
    {Institute of Estonian and General Linguistics, University of Tartu, Estonia},
    {Melbourne School of Psychological Sciences, University of Melbourne, Australia}
}

\leftheader{Aeschbach}

\abstract{Free-association norms provide essential empirical data for investigating linguistic, semantic, and cultural phenomena in the cognitive sciences. Although large-scale norms exist for languages such as English, Dutch, Spanish, and Mandarin Chinese, no comparable resource has been available for German. To address this gap, we present free-association norms for 5,877 German cue words as part of the German version of the multilingual Small World of Words (SWOW) project. We describe the data collection procedures, participant characteristics, and our comprehensive preprocessing pipeline before introducing the resulting SWOW-DE data set. Using data from three established psycholinguistic paradigms, we show that SWOW-DE norms robustly predict performance in lexical decision tasks, relatedness judgments, and psycholinguistic word ratings. Furthermore, we demonstrate that SWOW-DE responses compare favorably with existing German resources and provide a preliminary cross-linguistic comparison revealing both shared and language-specific association patterns, highlighting promising directions for future research. Overall, SWOW-DE represents the largest collection of German free associations to date and offers a unique resource for linguistic, psychological, and cross-cultural research.}

\authornote{
   \addORCIDlink{Samuel Aeschbach}{0000-0002-6167-4901},
   \addORCIDlink{Rui Mata}{0000-0002-1679-906X},
   \addORCIDlink{Kaidi Lõo}{0000-0002-4536-5287}
   \addORCIDlink{Simon De Deyne}{0000-0002-7899-6210},
   \addORCIDlink{Dirk U. Wulff}{0000-0002-4008-8022},

  Correspondence concerning this article should be addressed to Samuel Aeschbach, University of Basel, Faculty of Psychology, Center for Cognitive and Decision Sciences, Missionsstrasse 60/62, 4055 Basel, Switzerland, email: samuel.aeschbach@unibas.ch, and Dirk U. Wulff, Max Planck Institute for Human Development, Lentzeallee 94, 14195 Berlin, Germany, email: wulff@mpib-berlin.mpg.de}

\begin{document}
\maketitle

Free association has served as a window into human thought since the earliest days of psychology \parencite{galton_inquiries_1883, jung_association_1910}. In contemporary cognitive science, it has become a central tool for probing linguistic structure \parencite[][]{brochhagen_language_2023}, mapping the organization and dynamics of semantic memory \parencite{steyvers_large-scale_2005, wulff_new_2019}, studying (second) language learning \parencite{hills_longitudinal_2009, fitzpatrick_word_2020}, and examining cross-cultural variation in conceptual representations \parencite{wulff_semantic_2022}. Progress in these areas increasingly relies on large, high-quality free-association norms, which support the generation of empirically grounded experimental stimuli and simulations \parencite[e.g.,][]{herault_creative_2024, aeschbach_measuring_2025}, enable comparisons of semantic representations across individuals and populations \parencite[e.g.,][]{dubossarsky_quantifying_2017, wulff_semantic_2022}, and provide benchmarks for evaluating the alignment between humans and machines \parencite[e.g.,][]{abramski_llm_2025, hussain_probing_2024, de_deyne_predicting_2017, de_deyne_evaluating_2024}. 

Although several free-association norm data sets exist \parencite[e.g., ][]{kiss_associative_1973, nelson_university_2004}, like many foundational resources in psychology and linguistics, they remain concentrated on a small set of languages \parencite[e.g., ][]{bates_psycholinguistics_2001}. Given the language-specific nature of linguistic units, semantic memory, and conceptual structure, there is a clear need for parallel resources across more diverse linguistic populations \parencite{berghoff_diversity_2025}. German is an especially important case: As an official language in several European countries and spoken by an estimated 140 million people worldwide \parencite{eberhard_german_2026, eberhard_german_2026-1}, it represents a sizable and widely studied linguistic population. However, existing German free-association data sets \parencite{russell_complete_1970, schulte_im_walde_empirical_2008, schulte_im_walde_association_2015, jonauskaite_free_2025, wulff_data_2022} are relatively small in scale, covering at most a few hundred cue words based on limited participant samples.

The Small World of Words (SWOW) project (https://smallworldofwords.org/) addresses this need by collecting extensive, methodologically comparable free-association norms across multiple languages, including German. Previously published SWOW data sets include Dutch \parencite{de_deyne_better_2013}, English \parencite{de_deyne_small_2019}, Rioplatense Spanish \parencite{cabana_small_2023}, Mandarin Chinese \parencite{li_large-scale_2024}, and Slovene \parencite{vintar_swow-sl_2025, brglez_word_2024}. In the present work, we expand this multilingual effort by introducing the German Small World of Words data set (SWOW-DE).

We introduce the largest resource of German free-association norms to date. SWOW-DE comprises 5,877 cue words, offering a unique resource for psychological, linguistic, and computational research into German-speaking populations and in cross-linguistic comparison. In the following sections, we describe the methods used to collect and process the raw free associations, provide an overview of participant characteristics and the resulting data, and demonstrate the utility of SWOW-DE in two major domains: (a) cross-linguistic comparisons of free associations and semantic network structure, and (b) predictions of lexical decision times, word-relatedness judgments, and psycholinguistic word properties relative to fastText, an established text-based language model available in German \parencite{grave_learning_2018}. We conclude by discussing the implications of this data set for the field and outlining directions for future research.

\section{Method}

\subsection{Materials and Procedure}

The free-association data reported on here were collected as part of the multilingual Small World of Words (SWOW) online citizen science project. When potential participants visited the study website, they were presented with a short description of the SWOW project that motivates free-association research. Fluency in German is stated as a condition for participation, and links to versions of the SWOW project in other languages are provided. Upon entering the study and agreeing to the use of their anonymous data for research purposes, participants first provided demographic information, including their age, gender, native language, and education, and optionally their approximate geographic location. Participants then entered the free-association task, described in more detail in the next paragraph. After completing the free-association task, participants were thanked for their contribution to the project and, since the introduction of the feature in 2024, were offered the opportunity to compare their own associations to the most frequent ones for the same cues.

\subsubsection{Free-Association Task}

\begin{figure}
    \centering
    \includegraphics[width=0.9\linewidth]{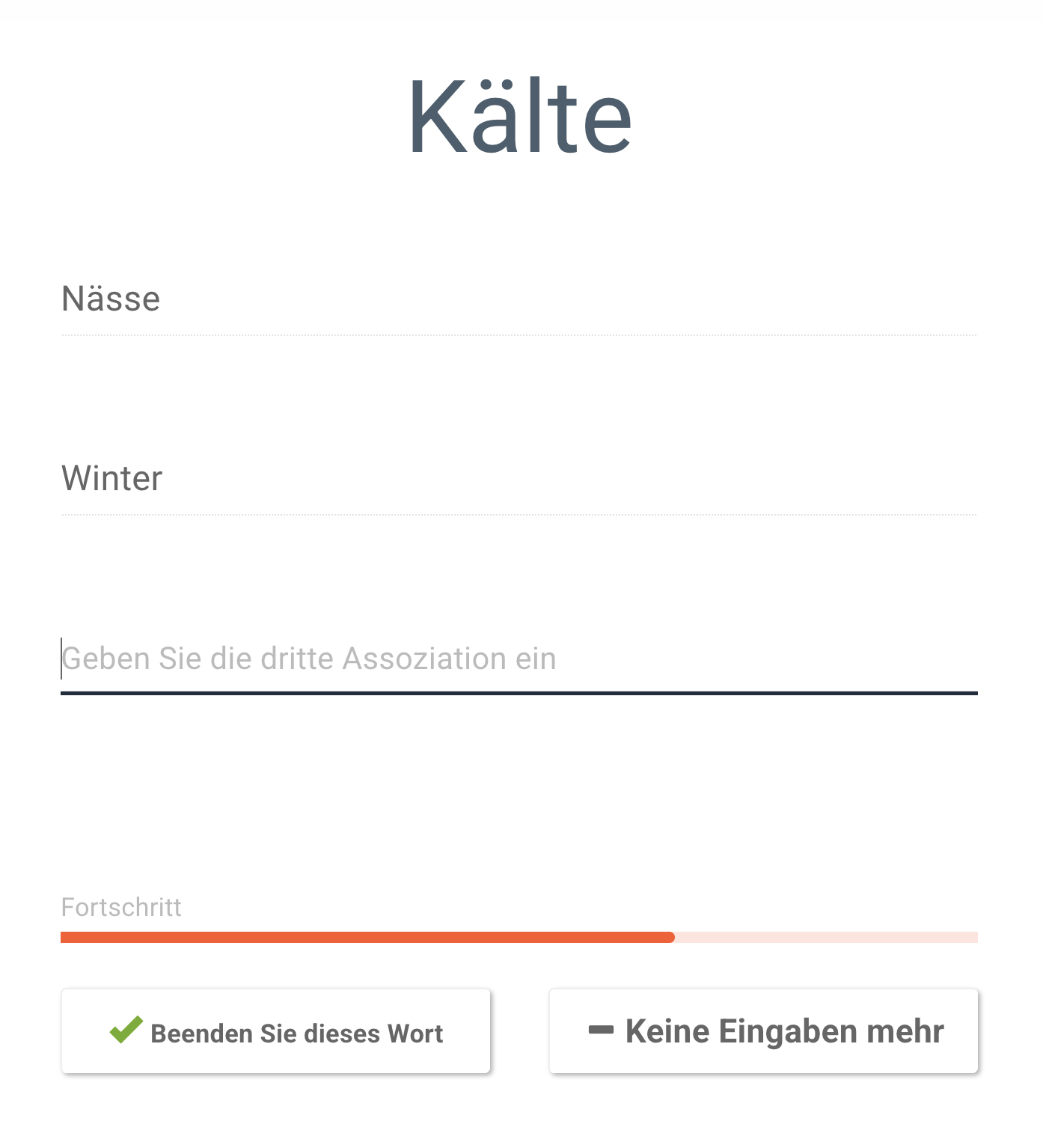}
    \caption{\textbf{Free-association task.} One trial showing the cue word ``Kälte'' (\textit{cold}) in the top center, and three text fields below for free-association responses. In the screenshot, the first two, ``Nässe'' (\textit{wettness}) and ``Winter'' (\textit{winter}), have already been filled out by the participant, the third is to be filled out next. Shown below are a progress bar and buttons to enter the next word or move on without a further response.}
    \label{fig:trial}
\end{figure}

Participants first saw a brief instruction about the free-association task, prompting them to respond with the first three words that came to mind when thinking about the cue word displayed at the top of the screen. Appendix~\ref{app:instruction} provides the full German instructions and an English translation. Figure~\ref{fig:trial} shows one trial of the free-association task, presenting a cue word in written form at the top of the screen, along with three text fields to enter associations. If a cue word was unknown to the participant, that could be indicated by clicking an appropriate button. After entering an association, it remained on screen in gray color until the next trial began with a new cue word. If a participant was unable to produce three associations, this could be indicated by clicking a button. Participants completed 14 to 18 trials, with each trial presenting a different cue word. Although the number of trials per participant and the visual appearance of the user interface evolved slightly throughout the project, the free-association task and its functions remained unchanged. 

\subsubsection{Cue Words}

\begin{figure*}
    \includegraphics[width=\textwidth]{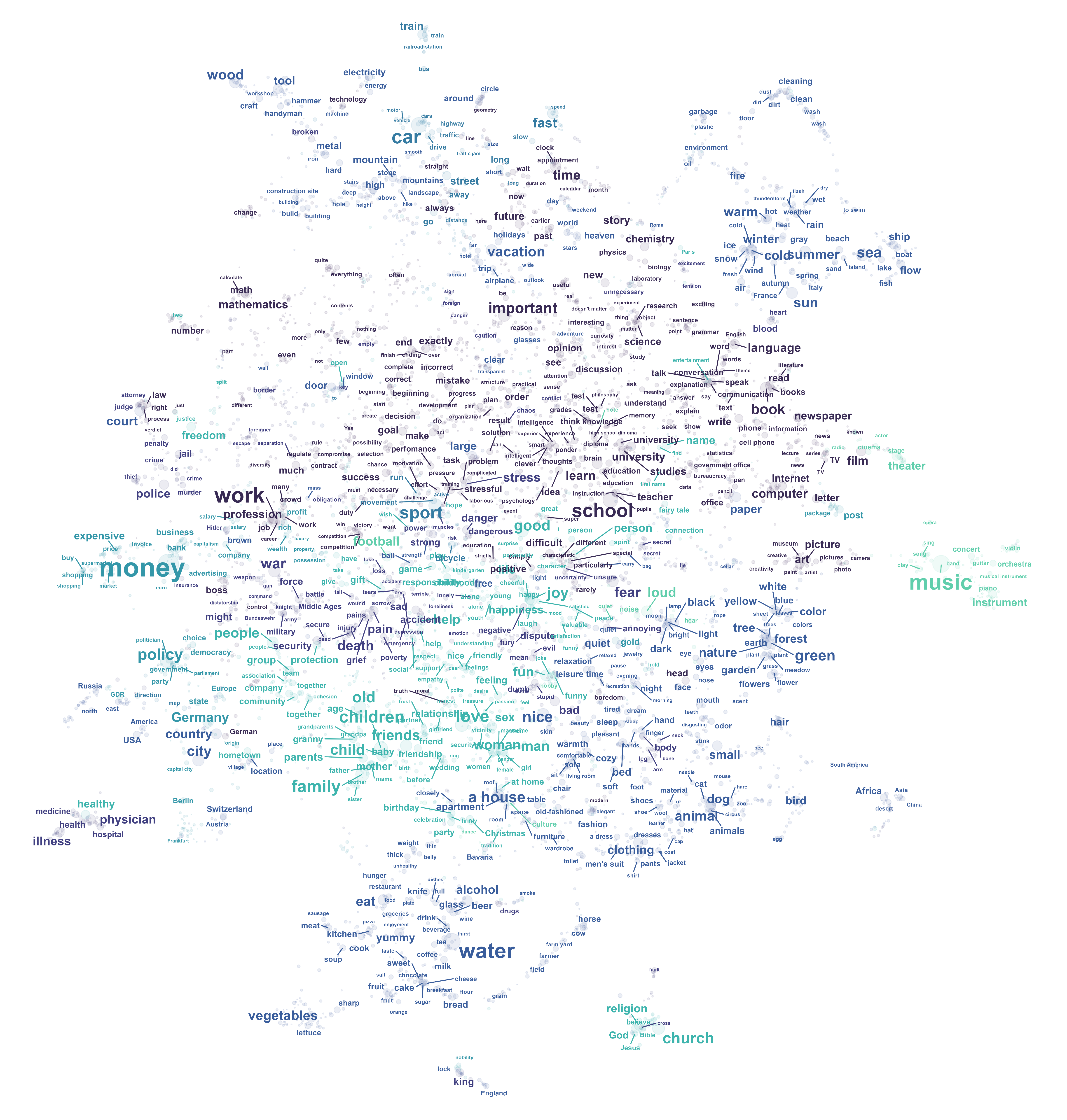}
    \caption{\textbf{Cue word set illustration.} The figure shows a uniform manifold approximation and projection (UMAP) \parencite{mcinnes_umap_2020} of cue words based on cue word embeddings derived from the positive pointwise mutual information (PPMI)-weighted co-occurrence matrix including all responses given at least five times, reduced via singular value decomposition (SVD) (see Figure~\ref{fig:transformations}D) using the associatoR package for R \parencite{aeschbach_mapping_2025}. Color indicates clusters found using the Louvain algorithm \parencite{blondel_fast_2008}. Only the top 15\% most frequent cue words among responses are labeled; larger points and labels indicate cue words more frequent among responses.}
    \label{fig:embedding}
\end{figure*}

The cue set was expanded iteratively throughout data collection. Cue words were initially selected based on high frequency in language, use as stimuli in other studies, part of speech, and, with increasing progress of the study, to include frequent responses to previous cues. Starting with a set of 1,036 cue words in 2012, sets of 1,642, 1,741, and 1,592 words were added whenever previous sets reached responses to each cue from at least 60 participants. Eight cues were excluded from the data set due to typographical errors and duplications, leading to an initial total set of 6,003 cue words. The final set of cue words in SWOW-DE is still smaller due to the exclusion of cues with too few responses, see section \textit{Data Filtering and Preprocessing}. However, the full data are also being made available (see \textit{Availability of Data and Materials}).

Figure~\ref{fig:embedding} presents a two-dimensional illustration of the final SWOW-DE cue words organized by semantic similarity, with colors indicating clusters and larger points and label font size indicating higher frequency of the cue word among the responses. Cue word labels were machine-translated into English for this figure; Figure~\ref{fig:embedding-de} shows the same illustration with the original German labels.

\subsection{Data Collection}

Data collection for SWOW-DE took place over more than a decade, from 2012 to 2025. The website running the study is publicly accessible at \href{https://smallworldofwords.org/de}{https://smallworldofwords.org/de}. The study was advertised through various channels, including institutional press releases, radio interviews, newspaper articles, advertising on public transport in multiple German cities, social media advertising, newsletters, university lectures, and mailing lists. Participation was fully voluntary and did not require any form of registration. Participants were not financially compensated; however, a small portion received course credit. 

Overall, these data collection efforts yielded a raw sample of 23,999 participants by August 25, 2025, the cut-off date for the data set shared with this article. Note that, because data collection was anonymous, participants could complete the task more than once, meaning the number of unique individuals is likely somewhat smaller.

\subsection{Data Filtering and Preprocessing}

The open nature of the data collection, the free-text responses, and the intricacies of German orthography, including regional variations, warranted a multistep filtering and preprocessing pipeline. This involved excluding participants who provided lower-quality responses, checking and correcting response spelling, and setting a minimum criterion for the number of trials per cue.

We applied a set of filtering criteria to exclude participants with too many invalid responses from the data. These criteria are similar to those employed in other SWOW projects. First, we excluded participants whose responses contained more than one word in more than 30\% of their responses (2.20\% of participants). Second, we excluded participants whose responses contained more than 20\% repeated responses (1.10\% of participants). Third, we excluded participants whose responses passed an initial spell-check in fewer than 60\% of cases (2.13\% of participants). And fourth, we excluded participants whose responses were ``unknown word'' or missing in more than 60\% of cases (5.97\% of participants). In total, 22,047 (91.87\%) participants, representing 367,454 (92.25\%) trials, fulfilled all four criteria. Appendix~\ref{app:participant_sel} provides additional details on participant inclusion.

To address misspellings and orthographic variation, we applied standard text data cleaning procedures, including the removal of most special characters and extra whitespace, and spellchecks in our data preprocessing pipeline, initially whitelisting 84.47\% of responses using an implementation of the Hunspell spell checker for R \parencite{ooms_hunspell_2025, hunspell_hunspell_2022}. To improve the remaining responses, we implemented three main steps. First, we normalized the use of umlauts (ä, ö, ü), which are often written as ``ae'', ``oe'', and ``ue'', of ``Eszett'' (ß) and double ``s'', which are often used inconsistently, and improved the casing of responses, which in German is more complex than in other languages covered by different SWOW projects. This was implemented by testing whether any of the possible combinations of these changes led to a correct German word. This stage also unified orthographic variations across German-speaking countries to the spelling rules valid in Germany in 2025 and increased whitelisted responses to 94.36\%. Second, as the remaining non-whitelisted responses included many proper nouns, such as movie titles, consumer brands, and public figures that Hunspell did not recognize, we evaluated their validity in the following way: For each response, we tested whether there is a German Wikipedia article with the response as a title, or a Wikipedia article with a related, possibly slightly differently spelled, article title that the Wikipedia's search API directs to when using the response as the search term. Whitelisting all responses for which a German Wikipedia article can be found and using the related article titles as spelling improvements increased whitelisted responses to 97.17\%. 

To evaluate and improve the spelling of the remaining 27,959 non-whitelisted responses, we employed open-weight, locally hosted large language models (LLM), specifically Meta's ``Llama 4 Scout (17Bx16E)'' \parencite{meta_llama_llama-4-scout-17b-16e-instruct_2025} and OpenAI's ``gpt-oss-120b'' \parencite{openai_gpt-oss-120b_2025}. We prompted the LLMs to return a correctly spelled version of the potentially misspelled response, given the context of the cue word and the other two responses provided by the same participant in the same trial. Appendix~\ref{app:prompts} provides a full report of the prompts used and model settings. To evaluate the model and prompt, we created a test set of human-spelling improvements for 1,000 cases of non-whitelisted, potentially misspelled responses: 500 cases for model and prompt selection, and 500 cases to evaluate model performance. The better model of the two, ``gpt-oss-120b'', showed an accuracy of $.66$. Applying the LLM-correction to the remaining 2.83\% responses further increased the whitelisted responses to a final 98.79\%. Note that SWOW-DE includes the raw responses, as well as a variable indicating the stage at which a response was corrected, allowing users to revert LLM-based or other corrections.

Finally, we set a minimum criterion for the number of trials per cue, which is important for several analyses, as otherwise cue representations or statistics would be informed by uneven amounts of information. We set a minimum of 55 responses, maximizing the minimum criterion and data retention. This led to the exclusion of 126 cue words (2.10\%) for which fewer than 55 response trials were available. For the remaining 5,877 cue words, we randomly sampled 55 response trials where more were available. In total, these two steps---the removal of cues with few responses and extra trials from cues with more than 55 responses---led to the exclusion of 44,219 trials (12.03\%) and the removal of 21 participants (0.10\%) from contributing any trials. Note that a version of the data, including all cues and all responses collected, is also available (see \textit{Availability of Data and Materials}). 

\section{Characterizing SWOW-DE}

In the following, we characterize SWOW-DE by analyzing participant demographics and basic cue and response statistics. 

\subsection{Demographic Characteristics}

\begin{figure*}
    \includegraphics[width=\textwidth]{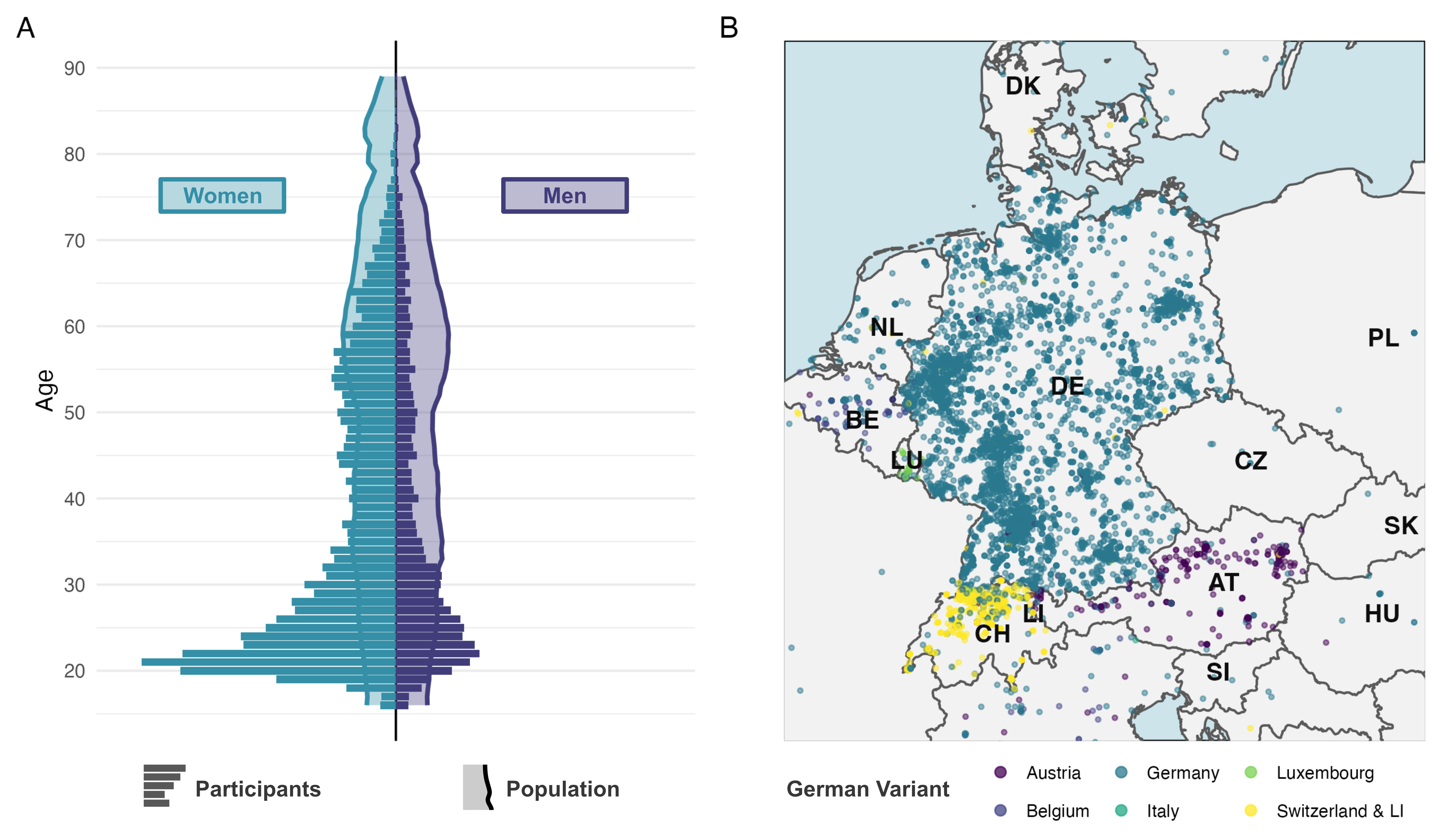}
    \caption{\textbf{Participant demographics.} (A) Participants' age distribution (horizontal bars) compared to the population of Germany in 2023 (lines and area) by gender. Note that population data are only available up to 89 years of age. (B) Approximate participant location and native German variant of participants included in SWOW-DE.}
    \label{fig:demographics}
\end{figure*}

Of the 22,026 participants in SWOW-DE, 15,590 (70.78\%) indicated ``female'' gender, 6,212 (28.20\%) indicated ``male'' gender, and 224 (1.02\%) indicated another, not further specified, gender labeled ``X''. The option to indicate a non-binary gender was not present at the start of data collection but was added at a later point. Participants were between 16 and 91 years of age ($M = 37.03, SD = 16.03$) at the time of participation. Figure~\ref{fig:demographics}A presents the participants' age distribution compared with demographic data for Germany \parencite{statistisches_bundesamt_german_federal_statistical_office_bevolkerung_nodate}, separately for men and women. Although SWOW-DE includes contributions from a broader set of individuals than most empirical studies, women and younger participants are overrepresented---a pattern also found in other SWOW samples.

The native language of participants was predominantly German (n = 20,701; 93.98\%). The majority of the participants (n = 16,606; 75.39\%) indicated ``German (Germany)'' as native language, followed by German variants from Switzerland and Liechtenstein (n = 3,222; 14.63\%), Austria (n = 723; 3.28\%), Belgium (n = 94; 0.43\%), Luxembourg (n = 47; 0.21\%), and Italy (n = 9; 0.04\%). These proportions broadly align with those of the German-speaking populations in these countries, though the Swiss/Liechtenstein variant is somewhat overrepresented. Figure~\ref{fig:demographics}B shows the approximate geographical locations reported by some of the participants. The reported native languages largely match the predominant German variant spoken in each region, and the spatial distribution of participants broadly tracks geographic population densities.

Information on participants' education level is available for 16,385 participants (74.39\%). Of these participants, 10,589 (64.63\%) report a higher education degree, 4,613 (28.15\%) a high school diploma, 1,049 (6.40\%) a secondary school diploma, 68 (0.42\%) elementary school only, and 66 (0.40\%) no formal degree. The sample is thus highly educated, more so than the general populations of German-speaking countries \parencite[cf. 33\% of adults aged above 15 years with a higher education degree in Germany, 43.8\% of working adults with higher education degree in Switzerland, and 21\% of adults aged 25--64 years with higher education degree in Austria;][]{bach_bildungsniveau_2024, bundesamt_fur_statistik_swiss_federal_statistical_office_bildungsniveau_2025, statistik_austria_statistics_austria_bildungsstand_2025}. 

\subsection{Cue Part-of-Speech, Response Types and Tokens}

The 5,877 SWOW-DE cue words consist of 3,921 nouns, 870 adjectives, 700 verbs, 292 adverbs, and 112 other parts of speech, according to an implementation of the Stanford Log-linear Part-of-Speech Tagger \parencite{the_stanford_natural_language_processing_group_stanford_nodate, schramm_wortarteninfo_nodate}. 

SWOW-DE comprises 868,814 responses, or \textit{tokens}; 318,104 in first, 291,914 in second, and 258,796 in third response position. These text responses contain 85,227 unique responses, or \textit{types}; 49,002 of which (57.50\%) appear only once in the data set. Such one-off types, although a considerable part of all types, only amount to 5.64\% of all tokens. These fractions are comparable to other SWOW data sets. For example, in the English data set, 57.3\% of types and 2.3\% of tokens appeared only once \parencite{de_deyne_small_2019}.

Among all responses, 66.76\% (580,010 tokens) are themselves cue words in SWOW-DE, which we refer to as \textit{coverage}. This coverage is somewhat lower than in other, larger SWOW data sets---the English data set, for instance, has a coverage of 87\% \parencite{de_deyne_small_2019}---but is sufficient to support important analyses, especially concerning the construction of networks and embeddings.

\section{Demonstrating Utility in Applications of Free Associations}

\begin{figure*}
    \includegraphics[width=\textwidth]{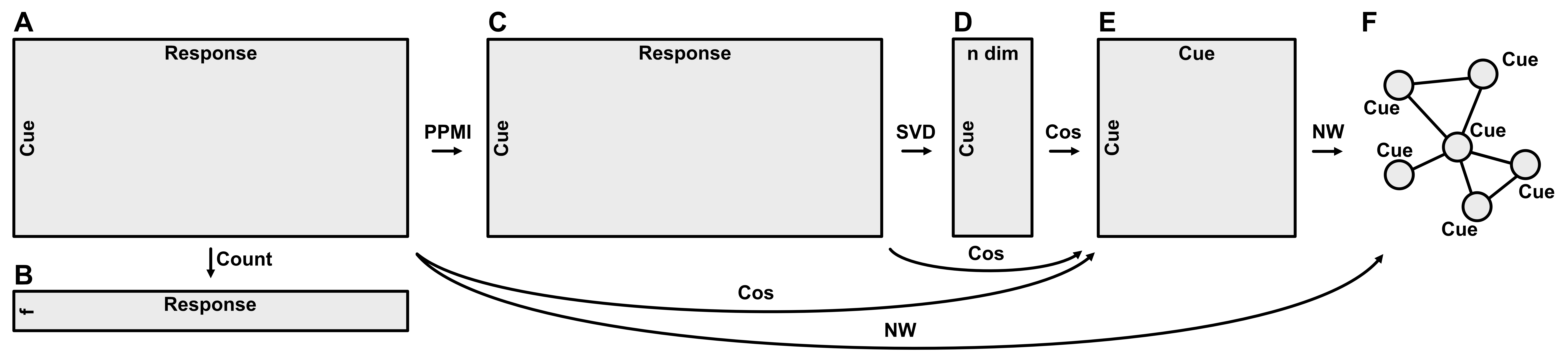}
    \caption{\textbf{Overview of transformations applied to free-association data.} (A) Raw cue--response frequency matrix, with cues as rows and responses as columns; each cell contains the number of times a response was produced for a cue. This standard representation can predict word similarity using response covariance. (B) Summary vector derived from (A), aggregating response frequencies across all cues, with responses as columns and total frequencies as values. This representation is used to predict response times in lexical decision tasks and to compare frequent responses across languages. (C) Matrix after applying a positive pointwise mutual information (PPMI) transformation to A, emphasizing informative cue--response associations, which improves word similarity predictions over raw counts. (D) Dimensionality-reduced cue representation obtained via singular value decomposition (SVD); cues are rows, latent dimensions are columns, and values indicate each cue's magnitude on a dimension. This representation further improves on the predictiveness of word similarity and is well-suited to predict word-level features, such as ratings on arousal or valence. (E) Cue--cue similarity matrix computed from representations in (A), (B), or (C) using a similarity function (here, cosine similarity; Cos). (F) Network representation inferred (NW) from cue--cue similarities in (E) or from cue--response associations in (A), with cues as nodes and association strengths as edges.}
    \label{fig:transformations}
\end{figure*}

In this section, we demonstrate the utility of SWOW-DE in two sets of analyses: comparing SWOW-DE to other languages and using SWOW-DE to predict psycholinguistic experimental results. These analyses showcase several transformations of SWOW-DE that produce representations emphasizing different properties of the associations and enabling different applications. Figure~\ref{fig:transformations} provides an overview of these transformations and representations, to which we will refer in the upcoming demonstrations. These analyses also showcase the performance users can expect, relative to alternative approaches (e.g., text-based word embeddings) and to SWOW data sets in other languages, when employing SWOW-DE in psycholinguistic and cognitive research.

A starting point for all transformations and analyses is the cue response matrix shown in Panel~A in Figure~\ref{fig:transformations}. It tabulates the frequency of response types (columns) across all cues (rows). For SWOW-DE, this matrix comprises 5,877 rows and 85,227 columns.  

\subsection{Comparing Free Associations Across Languages}

The SWOW project collects free-association data in 20 languages across multiple continents using a shared study design, enabling larger-scale cross-linguistic comparisons of associative patterns and semantic structure than were previously possible. Here, we compare SWOW-DE with publicly available data from other language projects, analyzing the most frequent responses and the semantic network structure in each language.

\subsubsection{Comparison of Response Frequencies}

\begin{table*}[htbp]
\vspace*{2em}
  \begin{threeparttable}
      \caption{Top 10 Free-Association Responses in SWOW Datasets}
    \label{tab:top10}
\begin{tabular}{rllllll}
\hline
\multicolumn{1}{c}{R} & \multicolumn{1}{c}{German} & \multicolumn{1}{c}{English\tabfnm{a}} & \multicolumn{1}{c}{Dutch\tabfnm{b}} & \multicolumn{1}{c}{Rioplatense Spanish\tabfnm{c}} & \multicolumn{1}{c}{Slovene\tabfnm{d}} & \multicolumn{1}{c}{Mandarin Chinese\tabfnm{e}}\\
\hline
1 & Geld \textit{(money)} & money & water \textit{(water)} & agua \textit{(water)} & denar \textit{(money)} & \begin{CJK}{UTF8}{gbsn}人\end{CJK} \textit{(person)}\\
2 & Musik \textit{(music)} & food & geld \textit{(money)} & comida \textit{(food)} & delo \textit{(work)} & \begin{CJK}{UTF8}{gbsn}钱\end{CJK} \textit{(money)}\\
3 & Arbeit \textit{(work)} & water & eten \textit{(food)} & amor \textit{(love)} & avto \textit{(car)} & \begin{CJK}{UTF8}{gbsn}水\end{CJK} \textit{(water)}\\
4 & Schule \textit{(school)} & car & pijn \textit{(pain)} & trabajo \textit{(work)} & šola \textit{(school)} & \begin{CJK}{UTF8}{gbsn}工作\end{CJK} \textit{(work)}\\
5 & Wasser \textit{(water)} & music & auto \textit{(car)} & dolor \textit{(pain)} & šport \textit{(sport)} & \begin{CJK}{UTF8}{gbsn}可爱\end{CJK} \textit{(cute)}\\
6 & Essen \textit{(food)} & green & lekker \textit{(tasty)} & dinero \textit{(money)} & služba \textit{(job)} & \begin{CJK}{UTF8}{gbsn}红色\end{CJK} \textit{(red)}\\
7 & Auto \textit{(car)} & red & muziek \textit{(music)} & música \textit{(music)} & knjiga \textit{(book)} & \begin{CJK}{UTF8}{gbsn}老师\end{CJK} \textit{(teacher)}\\
8 & Liebe \textit{(love)} & love & mooi \textit{(beautiful)} & animal \textit{(animal)} & čas \textit{(time)} & \begin{CJK}{UTF8}{gbsn}游戏\end{CJK} \textit{(game)}\\
9 & alt \textit{(old)} & work & kinderen \textit{(children)} & vida \textit{(life)} & človek \textit{(person)} & \begin{CJK}{UTF8}{gbsn}时间\end{CJK} \textit{(time)}\\
10 & Familie \textit{(family)} & old & school \textit{(school)} & casa \textit{(house)} & ljubezen \textit{(love)} & \begin{CJK}{UTF8}{gbsn}朋友\end{CJK} \textit{(friend)}\\
\hline
\end{tabular}
\vspace*{0.5em}
    \begin{tablenotes}[para,flushleft]
        {\small
            \textit{Note.} Top 10 responses of SWOW-DE-2025-R55 cue words compared to English, Dutch, Rioplatense Spanish, Slovene, and Mandarin Chinese SWOW datasets. R: Rank.
            \tabfnt{a}\textcite{de_deyne_small_2019}
            \tabfnt{b}\textcite{de_deyne_better_2013}
            \tabfnt{c}\textcite{cabana_small_2023}
            \tabfnt{d}\textcite{brglez_word_2024}
            \tabfnt{e}\textcite{li_large-scale_2024}
         }
    \end{tablenotes}
\end{threeparttable}
\end{table*}

Response frequency in free-association tasks can be understood as a measure of cultural importance. We evaluated the most frequent responses in SWOW-DE against all publicly available SWOW data sets \parencite[in English, Dutch, Rioplatense Spanish, Mandarin Chinese, and Slovene; ][]{de_deyne_better_2013, de_deyne_small_2019, cabana_small_2023, li_large-scale_2024, brglez_word_2024}, by summing the frequency of response types across all cues (Panel B in Figure~\ref{fig:transformations}). Table~\ref{tab:top10} shows the 10 most frequent words in each language. Despite geographic and cultural differences, the most frequent responses largely overlapped. Responses such as money, water, food, work, and love are top responses across many of the languages; however, the ordering and entries of the top 10 responses vary. In the German top 10, for instance, the second most frequent response, music, is listed much higher than in other languages, whereas the fourth most frequent response, school, appears only in two other languages. More systematic analyses of word-frequency differences could uncover general trends across languages and possible culture and language-specific differences.

\subsubsection{Comparison of Semantic Networks}

Semantic networks capture the structure of relationships among words and can be characterized in terms of their connectedness, structure, and efficiency \parencite{siew_cognitive_2019, borge-holthoefer_semantic_2010, kumar_critical_2022}. Semantic networks have been used to model the structural underpinnings of semantic cognition \parencite{steyvers_large-scale_2005}, and elucidate cognitive differences across individuals \parencite{morais_mapping_2013}, and people of different ages \parencite{dubossarsky_quantifying_2017, wulff_using_2022} or creativity \parencite{benedek_how_2017, kenett_investigating_2014}. Here, we use semantic networks to analyze differences in the semantic organization of languages.

There are several methodological approaches for constructing semantic networks from free-association data (see Figure \ref{fig:transformations}). We chose a direct approach by constructing a cue--cue network with directed edges reflecting the number of times that a cue produced other cues as responses (direct path from Panel~A in Figure~\ref{fig:transformations}). This resulted in a directed network with 5,877 nodes and 242,330 edges (Panel~F in Figure~\ref{fig:transformations}). We used the \textit{igraph} package version 2.1.4 for R \parencite{csardi_igraph_2025} for all network calculations. 

Table~\ref{tab:networks_matched} shows the network measures for the matched German, Dutch, English, Rioplatense Spanish, and Mandarin Chinese semantic networks. Focusing on SWOW-DE, shown in the first row, we observe a small-world network structure \parencite{watts_collective_1998} characterized by high clustering compared to a weighted, directed random network of the same size ($Avg. CC_{SWOW-DE} = 0.14$, $Avg. CC_{Random} = 0.01$) and a short average path length compared to a weighted, directed regular lattice of similar size ($ASPL_{SWOW-DE} = 111.45$, $ASPL_{Regular} = 1683.06$). Small-world structure has been repeatedly observed for semantic networks \parencite{steyvers_large-scale_2005, wulff_structural_2022}, is a feature of many naturally occurring networks, and is thought to facilitate efficient information transmission \parencite{watts_collective_1998}. 

How does the structure of the SWOW-DE network compare to those of other languages? For meaningful comparisons, the size of the networks must be held constant; in the chosen approach, this is determined by the number of cues. As SWOW projects differ substantially in the number of cues that they include, we had to equate the networks across all languages. To do this, we selected the cue words most frequently mentioned in the responses in each language, matching the number of cues (5,877) in the German data, and sampled 55 response triplets per selected cue. This sampling approach reduced the networks to the most important 5,877 words in each data set and mirrors the cue word selection process in SWOW project expansions. Due to their much smaller size, the Slovene data \parencite{brglez_word_2024, vintar_swow-sl_2025} could not be equated in this way and were therefore excluded from this comparison.

\begin{table*}[htbp]
\vspace*{2em}
  \begin{threeparttable}
      \caption{Semantic Network Properties of Multiple Languages' SWOW Free Associations}
    \label{tab:networks_matched}
\begin{tabular}{lccccc}
\hline
\multicolumn{1}{l}{Dataset} & \multicolumn{1}{c}{Language} & \multicolumn{1}{c}{Number of Edges} & \multicolumn{1}{c}{Average Strength} & \multicolumn{1}{c}{ASPL\tabfnm{a}} & \multicolumn{1}{c}{Average CC\tabfnm{b}} \\ 
\hline
SWOW-DE & German & 242,330 & 197.38 & 111.45 & 0.14 \\
SWOW-NL & Dutch & 296,052 & 253.98 & 109.03 & 0.14 \\
SWOW-EN & English & 311,682 & 246.39 & 100.85 & 0.11 \\
SWOW-RP & Rioplatense Spanish & 272,948 & 202.88 & 108.71 & 0.13 \\
SWOW-ZH & Mandarin Chinese & 280,764 & 192.79 & 108.82 & 0.12 \\
\hline
\end{tabular}
\vspace*{0.5em}
    \begin{tablenotes}[para,flushleft]
        {\small
            \textit{Note.} Properties of semantic networks generated from the German, English, Rioplatense Spanish, and Mandarin Chiniese SWOW free-association data. Results of non-German networks are based on subsets matching the German data in numbers of cues and responses per cue. All semantic networks have 5,877 nodes. \tabfnt{a} weighted, directed \tabfnt{b} unweighted, directed
         }
    \end{tablenotes}
\end{threeparttable}
\end{table*}

Table~\ref{tab:networks_matched} shows the network measures for all included languages, revealing striking similarities: Number of edges, average strength, average shortest path-length (ASPL; directed, weighted), and average local clustering coefficient (Average CC; directed, unweighted) were all within 24\% of the largest value. These results show that despite geographic, cultural, and demographic differences, free-association language networks exhibit similar small-world-like structures, suggesting a strong universal component. We hope that these preliminary analyses motivate future investigations into the commonalities and differences between associations and semantic networks across languages, leveraging the unique opportunities offered by the international SWOW project. 

\subsection{Predicting Psycholinguistic Data using Free Associations}

Free association is a primary approach to estimating the representational underpinnings of semantic cognition \parencite[]{de_deyne_small_2019, wulff_using_2022} in the form of semantic networks, word embeddings, or response frequency distributions. Here, we show that these representations can be used to accurately predict key psycholinguistic measures, including reaction times in recognizing words among pseudo-words, judgments of the relatedness of word pairs, and ratings of word properties such as valence. 

\subsubsection{Predicting Lexical Decision Task Reaction Times using Response Frequencies}

The lexical decision task (LDT) tests word recognition by presenting participants with a mix of words and pseudo-words. Participants then decide, as quickly as possible, whether a presented string of letters is a word in a given language. It is widely established that response times in LDT are correlated with corpus-based word frequency \parencite[e.g.,][]{balota_visual_2006, gerhand_age_1999, brysbaert_word_2011}. Moreover, in other languages, free-association response frequencies have been found to match or exceed predictive accuracy of corpus-based word frequencies \parencite{de_deyne_better_2013, de_deyne_small_2019, cabana_small_2023, li_large-scale_2024}.

\begin{figure*}
    \centering
    \includegraphics[width=\textwidth]{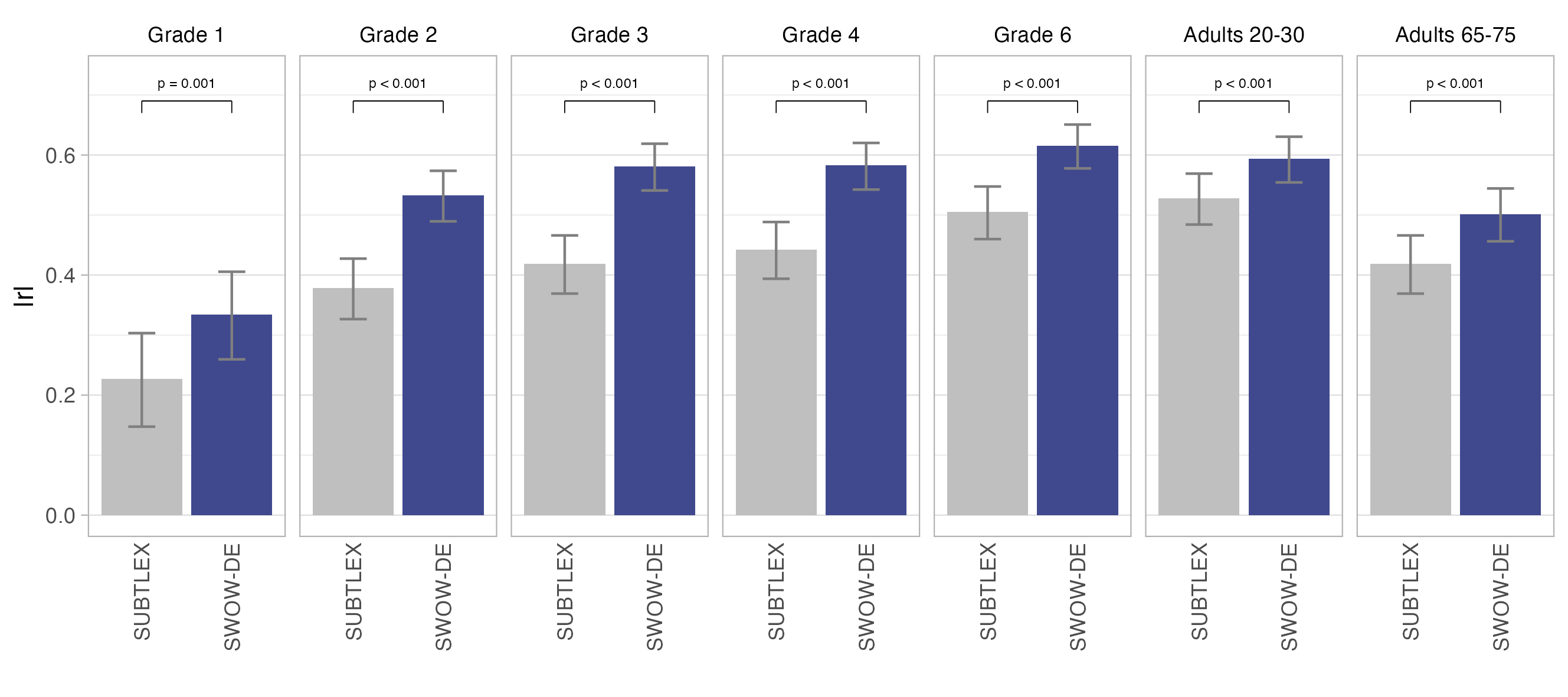}
    \caption{\textbf{Validation with lexical decision task response times.} Absolute Pearson correlation coefficients $\left| r \right|$ for LDT reaction times of 1,152 (576 for Grade 1) German words \parencite{schroter_developmental_2017} and SUBTLEX-DE word frequency \parencite{brysbaert_word_2011} as well as SWOW-DE free-association response frequency. Correlations are shown for each of the age groups in \textcite{schroter_developmental_2017}. SUBTLEX-DE and SWOW-DE frequency measures are log-transformed using base 10. Error bars indicate 95\% confidence intervals, p-values between SUBTLEX-DE and SWOW-DE correlations are based on $Z$ tests for equality of pairwise dependent correlations proposed by \textcite{steiger_tests_1980}.}
    \label{fig:validation-ldt}
\end{figure*}

Here, we test the predictive accuracy of free associations for LDT response times against corpus-based German word frequencies. As the benchmark, we use the ``DeveL'' data \parencite{schroter_developmental_2017} containing LDT responses to 1,152 (576 for Grade 1) German words and pseudo-words collected from a total of 782 German-speaking schoolchildren in Grades 1 to 6, as well as younger (20--30 years) and older (65--75 years) adults. To predict LDT response time, we aggregated response frequencies over all cues in SWOW-DE (cf. Panel B in Figure~\ref{fig:transformations}). As a comparison baseline, we used German SUBTLEX word frequencies derived from a German subtitle corpus \parencite{brysbaert_word_2011}. The SUBTLEX data have previously been shown to correlate more strongly with LDT response times than other available German word-frequency sources \parencite{brysbaert_word_2011}. Word frequencies from both SWOW-DE and SUBTLEX were log-transformed (base-10) before correlating them with LDT response time. We report the absolute Pearson correlation as a measure of predictive accuracy.

Figure~\ref{fig:validation-ldt} shows correlations ($\left| r \right|$) with LDT response times for each age group, separately for SUBTLEX and SWOW-DE. Except for very young schoolchildren (Grade 1), we observed high correlations between $.5$ and $.6$ for SWOW-DE, comparable to those found in analyses of English, Dutch, Rioplatense Spanish, and Mandarin Chinese SWOW data \parencite{de_deyne_better_2013, de_deyne_small_2019, cabana_small_2023, li_large-scale_2024}. These correlations were 12.48\% to 47.51\% higher than those observed for SUBTLEX \parencite{brysbaert_word_2011}; all comparisons $p\leq 0.001$ using the method proposed by \textcite{steiger_tests_1980}. Note that the German SUBTLEX seems to correlate less with LDT response time ($.22 < r < .53$) than SUBTLEX in other languages \parencite[e.g., $.55 < r < .65$ for English and Spanish SUBTLEX,][]{cabana_small_2023, de_deyne_small_2019}, which may be due to the differences between SUBTLEX variants or the German LDT data set employed in our analysis. Either way, SWOW-DE appears to provide the best available predictions of LDT data in German to date.

\subsubsection{Predicting Relatedness Judgments and Word-Ratings using Free Associations}

A particular form of representation that can be extracted from free associations is word embeddings \parencite[]{hussain_novel_2024, hussain_probing_2024, cabana_small_2023, li_large-scale_2024}. Word embeddings represent the meaning of words as high-dimensional numerical vectors, which can be used to evaluate semantic similarity between words or to directly decode semantic features, such as valence, using predictive models. Here, we explore the utility of SWOW-DE embeddings for predicting human word-relatedness judgments and psycholinguistic word ratings.

\begin{figure*}
    \centering
    \includegraphics[width=\textwidth]{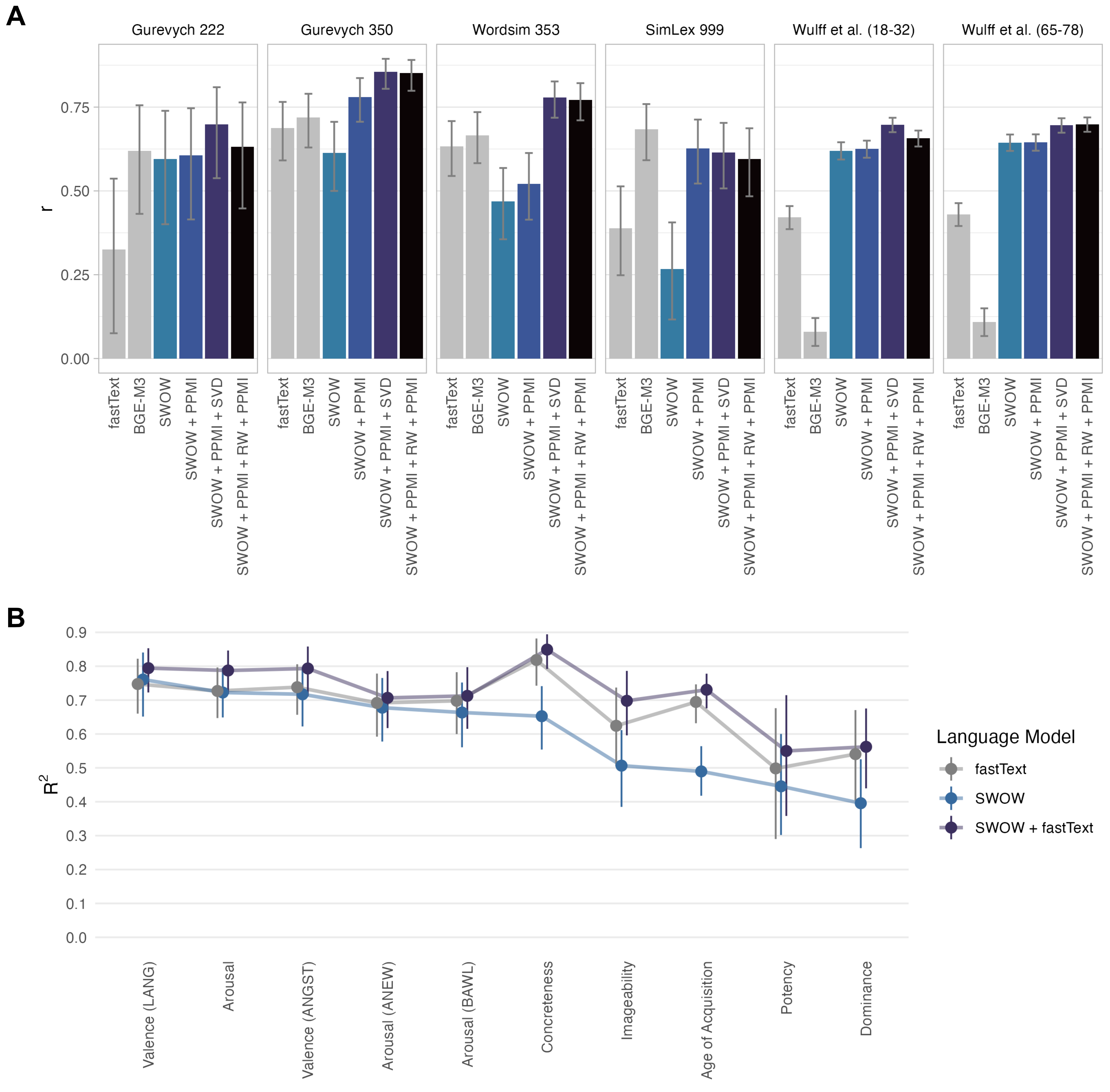}
    \caption{\textbf{Predictions of Relatedness Judgments and Psycholinguistic Word Attributes.} (A) Pearson correlation coefficients ($r$) between six data sets of empirical relatedness judgments and SWOW-DE-based relatedness (blue bars) as well as comparison (large) language model-based relatedness (gray bars). Error bars indicate 95\% confidence intervals of correlation coefficients. (B) Explained variance ($R^2$) of German word norms by three language models using 10-fold cross-validated ridge regression trained on 80\% of the data and evaluated on 20\% of the data. Ordered left to right by decreasing SWOW-DE $R^2$. Vertical lines indicate bootstrapped 95\% confidence intervals for $R^2$.}
    \label{fig:embedding_prediction}
\end{figure*}

\paragraph{Word Relatedness Judgments}

The relatedness judgment task presents participants with two words and asks them to judge their relatedness on a numerical rating scale \parencite[e.g.,][]{benedek_how_2017, kraemer_sequential_2021, wulff_structural_2022}. We compiled relatedness judgment data sets from several sources: 350 German word pairs from the Multilingual WS 353 data set\footnote{retrieved from \href{https://github.com/nmrksic/eval-multilingual-simlex/tree/master/evaluation/ws-353}{https://github.com/nmrksic/eval-multilingual-simlex/tree/master/evaluation/ws-353}} (\textit{Wordsim 353}) described in \textcite{leviant_separated_2015}, 999 German word pairs from an adaption of the SimLex-999\footnote{retrieved from \href{https://github.com/iraleviant/eval-multilingual-simlex/blob/master/evaluation/simlex-german.txt}{https://github.com/iraleviant/eval-multilingual-simlex/blob/master/evaluation/simlex-german.txt}} (\textit{SimLex 999}) described in \textcite{leviant_separated_2015}, two German-only relatedness judgment data sets with 222, and 350 words\footnote{retrieved from \href{https://tudatalib.ulb.tu-darmstadt.de/handle/tudatalib/2440}{https://tudatalib.ulb.tu-darmstadt.de/handle/tudatalib/2440}} (\textit{Gurevych 222}, \textit{Gurevych 350}) described in \textcite{gurevych_using_2005, gurevych_german_2020}, and two German-only relatedness judgment data sets on 63 animals from younger and older adults (\textit{Wulff et al. (18--32)}, \textit{Wulff et al., (65--78)}) described in \textcite{wulff_structural_2022}. Together, the data cover multiple parts of speech and words from diverse categories and levels of concreteness; however, they are somewhat biased towards concrete nouns, and, in the case of \textcite{wulff_structural_2022}, a specific category (animals).

To predict word-relatedness judgments, we extracted four word embeddings from SWOW-DE that represent different levels of inference. The baseline embedding is the matrix of $cues \times responses$ containing cue-specific response type frequencies (\textit{SWOW}; Panel~A in Figure~\ref{fig:transformations}). The second embedding applies a positive pointwise mutual information (PPMI) transformation that accounts for overall variability of cue and response type frequencies and has been shown to outperform the raw frequencies of the baseline embedding \parencite[e.g.,][]{salle_why_2019, bullinaria_extracting_2007} (\textit{SWOW + PPMI}; Panel~C in Figure~\ref{fig:transformations}). The third embedding applies singular value decomposition (SVD) to the PPMI-transformed frequencies to compress the column dimensionality to 300 dimensions and infer unobserved relationships between cues (\textit{SWOW + PPMI + SVD}; Panel~D in Figure~\ref{fig:transformations}). 
The fourth approach is technically not an embedding but instead estimates relatedness directly using a decaying random-walk process \parencite{abbott_random_2015}. Specifically, this approach first extracts a strongly connected component that includes all cues appearing in the responses and then applies PPMI, random-walk (RW) inference, and PPMI again (\textit{SWOW + PPMI + RW + PPMI}). For details on this approach, see \textcite{de_deyne_small_2019}.

As benchmarks for the free-association-based embeddings, we also obtained two text-based language models' embeddings; the German fastText model \parencite[\textit{fastText}; ][]{grave_learning_2018} and the multilingual BGE-M3 model \parencite[\textit{BGE-M3}; ][]{chen_bge_2024}. These models serve as representative examples of non-contextual (fastText) and contextual (BGE-M3) text-based embedding approaches. We extracted non-contextualized embeddings from BGE-M3 by inputting the words in isolation \parencite{tikhomirova_where_2026}. To generate predictions across all embedding models, we computed the cosine similarity between the embeddings of each word pair for each model.

Figure~\ref{fig:embedding_prediction}A shows Pearson correlation ($r$) between human-relatedness judgments and the five embedding models. SWOW-DE embeddings showed high correlations, with an average $\bar{r} = .66$ across data sets and embedding types. High-inference SWOW-DE embeddings based on singular value decomposition and random-walk inference performed best, with singular value decomposition achieving a slightly higher average performance of $\bar{r}_{SWOW + PPMI + SVD} = .73$ as compared to random-walk inference $\bar{r}_{SWOW + RW + PPMI} = .71$. The performance of both far exceeded that of the text-based embeddings ($\bar{r}_{BGE-M3} = .52$; $\bar{r}_{fastText} = .49$); however, the advantage of SWOW-DE varies considerably across data sets. These results are consistent with similar results for English, Rioplatense Spanish, and Mandarin Chinese SWOW data \parencite{de_deyne_small_2019, cabana_small_2023, li_large-scale_2024}. Together, they show that SWOW-DE permits accurate estimation of semantic relatedness, exceeding those of much larger text corpus-based language models, including modern transformer-based language models \parencite[for an overview, see][]{hussain_tutorial_2024}. 

\paragraph{Psycholinguistic Word Ratings}

Another application of embedding models is to decode from them features of the embedded words by means of predictive models. Here, we test how well SWOW-DE embeddings can be used to predict psycholinguistic word norms \parencite[e.g.,][]{hussain_probing_2024, vankrunkelsven_predicting_2018} relative to text-based models. 

We obtained word norms on valence \parencite{schmidtke_angst_2014, kanske_leipzig_2010}, dominance \parencite{schmidtke_angst_2014}, arousal \parencite{schmidtke_angst_2014, kanske_leipzig_2010}, concreteness \parencite{kanske_leipzig_2010}, age of acquisition \parencite{birchenough_rated_2017}, imageability \parencite{schmidtke_angst_2014}, and potency \parencite{schmidtke_angst_2014}. Each source included word ratings for about 1,000 German words, with age-of-acquisition ratings available for 3,259 words. For each word, we then extracted embedding vectors from the best-performing SWOW-DE-based and text-based language models (\textit{SWOW + PPMI + SVD} and \textit{fastText}). Next, we L2-normalized the vectors in each embedding to align their lengths, and then predicted each word rating from the z-standardized embedding dimensions using ridge regression; we evaluated performance using $R^2$ values obtained from predicting a held-out set of test data (20\%) using a model fitted on training data (80\%) with 10-fold cross-validation \parencite[for a similar approach, see][]{hussain_probing_2024, hussain_novel_2024}. In addition to evaluating each embedding model separately, we also tested a concatenation of the models to evaluate whether SWOW-DE can add to the predictive accuracy of the text-based model \parencite[see also, ][]{hussain_probing_2024}. Note that we extracted 300-dimensional embeddings from SWOW-DE, matching the dimensionality of fastText, so that both embeddings are weighted equally in the concatenated 600-dimensional embedding. 

Figure~\ref{fig:embedding_prediction}B shows the predictive performance by word norms ordered according to $R^2_{SWOW}$. Psycholinguistic word norms were overall accurately captured with an average $R^2 = .67$ and performances ranging from .40 to .85. The SWOW-DE $\bar{R^2}_{SWOW} = .60$ ($R^2_{SWOW} \in [.40, .76]$) embedding performed overall slightly worse than the fastText embeddings $\bar{R^2}_{fT} = .67$ ($R^2_{fT} \in [.50, .82]$); however, this difference appears to be driven by a subset of norms, especially concreteness and age of acquisition. The concatenated model consistently outperformed each individual embedding with $\bar{R^2}_{SWOW + fT} = .72$ ($R^2_{SWOW + fT} \in [.55, .85]$). Together, these results show that SWOW-DE enables accurate decoding of key psycholinguistic variables at a level almost comparable to that of a text-based language model trained on orders of magnitude more data. Moreover, the fact that the concatenated embedding outperformed the individual models further suggests that SWOW-DE contains psycholinguistic information not encoded in the text-based model \parencite[cf][]{de_deyne_visual_2021, hussain_novel_2024}. These results coincide with other results highlighting the value of combining representations from different data sources \parencite[e.g.,][]{de_deyne_visual_2021, hussain_probing_2024}. Overall, our demonstrations show that SWOW-DE is a powerful resource that extends the German psycholinguistic toolkit by improving predictions of psycholinguistic data, both as a standalone resource and alongside other resources. 

\section{Discussion}

We presented the Small World of Words free-association norms for the German language. This ongoing, multifaceted data-collection effort has produced the largest set of German free-association norms to date. The resulting SWOW-DE data set comprises 5,877 German cue words, each presented to at least 55 respondents who provided three responses per cue, yielding just under one million free-association responses. Our open, web-based recruitment strategy led to a heterogeneous sample of German-speaking adults, spanning a wide age range and representing both urban and rural regions across multiple European countries. As a result, the data set not only broadens the linguistic and cultural diversity available in psychological research but also offers a rich representation of the German-speaking population.

We demonstrated the utility of our free-association data in multiple applications. First, we compared the German free associations with those in other languages in the Small World of Words project for which similar data sets using the same general methodology have been assessed. Initial comparisons of most frequent responses showed high similarity between German, English, Dutch, Rioplatense Spanish, Mandarin Chinese, and Slovene data, despite the large geographical and cultural distance among some of the compared populations. Specifically, free-association networks derived from cue–response relationships showed comparable structural properties across languages, including small-world organization, demonstrating that patterns observed in earlier SWOW data sets extend to German, and suggesting a universal structure of semantic networks from free associations.

Second, we demonstrated utility in predicting psycholinguistic empirical data on the word and word-pair level. This included showing that our free-association responses correlate more strongly with lexical decision response times than SUBTLEX word frequency \parencite{brysbaert_word_2011}. We next demonstrated that language models created from the cue-specific response frequencies in our data correlate highly with human-relatedness judgments of word pairs and outperform standard corpus-based language models \parencite[BGE-M3 and fastText; ][]{chen_bge_2024, grave_learning_2018}. Finally, we showed that a ridge regression based on a free-association embedding model results in a prediction of psycholinguistic word properties comparable to a standard corpus-based language model \parencite[fastText; ][]{grave_learning_2018}, with few exceptions. It is important to note that these results have been achieved using a considerably smaller data set than was available in previous analyses of SWOW data in other languages \parencite{de_deyne_better_2013, de_deyne_small_2019, cabana_small_2023, li_large-scale_2024}, and with a fraction of the data used to train text-based models. 

All in all, these applications highlight that SWOW-DE efficiently captures meaningful aspects of human behavior, establishing SWOW-DE as a multipurpose window into human semantic knowledge and cognition in general. 

\subsection{Implications}

SWOW-DE provides several benefits for psychological, linguistic, and related research. First, the data set offers a comprehensive set of German free-association norms that can be used directly in studies with German-speaking participants, eliminating the need to rely on translated norms or proxy estimates from other languages. Using language-specific norms reduces translation-related artifacts and better reflects the linguistic and experiential background of the population under study. Second, the availability of a large-scale German data set contributes to greater linguistic diversity in behavioral and psycholinguistic research, which remains dominated by evidence from a small number of languages \parencite{berghoff_diversity_2025}. Third, because SWOW-DE follows the same study design as other SWOW data sets, it facilitates systematic cross-linguistic analyses of free-association data, increasing the interpretability and scope of comparative work as additional data sets become available. Fourth, validation analyses show that SWOW-DE norms perform favorably compared to text-corpus-based language models in predicting empirical psycholinguistic data. Using SWOW-DE instead of alternative sources of word frequencies or embeddings may therefore improve predictions of human behavior. In addition, unlike text-corpus-based language models, SWOW-DE provides detailed information about its underlying data, including participant demographics, time of data collection, and location data, enabling transparent reporting of potential biases and, where necessary, targeted filtering by time period, participant characteristics, or geographic region.

\subsection{Limitations}

There are some important limitations to the data and analyses presented here. First, although SWOW-DE is the largest collection of German free associations to date, it is smaller than most other published SWOW data sets (e.g., English: 12,292 cues, 100 participants per cue; Dutch: 12,571 cues, 100 participants per cue). By prioritizing words that occur frequently in text corpora and association responses during cue set expansion, SWOW-DE optimizes usefulness for its size, and as the evaluation results demonstrate, its current scale is sufficient to match or exceed the performance of alternative models in many applications. Because data collection is ongoing, future releases will further expand the data set. Second, the long data-collection period and iterative cue set expansion introduce potential confounds, including period effects. We have not explicitly tested for such effects; however, SWOW-DE includes timing information that would allow such analyses in future work. Third, the cross-linguistic network comparisons required equating data set sizes by selecting the most frequent cues from larger SWOW projects and sampling 55 response trials per cue. Although this controlled for differences in word frequency, future research should explore more refined matching strategies, such as cue-by-cue, translation-based alignment. Finally, comparisons of predictive performance with other SWOW validation studies are potentially affected by smaller German LDT and relatedness judgment data sets than those available in other languages. Nonetheless, the replication of similar effects across languages and the common inclusion of SimLex data \parencite{hill_simlex-999_2015, leviant_separated_2015}, which was previously used to validate data sets in other languages, strengthen overall confidence in free-association norms as robust predictors of psycholinguistic outcomes.

\subsection{Future Work}

Several directions for future work follow from the present study. At the level of data collection, SWOW-DE could be expanded by increasing both the number of responses per cue and the number of cue words, bringing it closer in scope to larger SWOW data sets available for other languages. In addition, greater harmonization of cue selection across SWOW languages, for example, with respect to psycholinguistic dimensions or conceptual coverage, could further improve the comparability of data sets in cross-linguistic analyses. However, such efforts should be balanced with efforts to cover the most important words in each language.

Beyond data expansion, SWOW-DE invites a range of analytic extensions. These include assessing whether free-association-based models improve predictions of human behavior in additional experimental paradigms and exploring how alternative sampling or aggregation strategies affect derived representations. Further, examining subsets of SWOW-DE defined by participant characteristics or time periods may yield more accurate predictions for corresponding outcomes, providing insights into semantic structure and conceptual variation. The cross-linguistic comparisons presented here are intentionally limited in scope; more systematic comparative analyses across languages could help identify shared and language-specific patterns in free associations, with implications for theories of language processing and semantic memory. Finally, SWOW-DE offers opportunities to evaluate and better align text-corpus-based language models, including large language models, with human semantic behavior.

In sum, we introduced the SWOW-DE free-association data set, documented its collection and preprocessing, and demonstrated its utility across a range of applications, including cross-linguistic comparisons and the prediction of empirical psycholinguistic data. Together, these results establish SWOW-DE as a robust, transparent, and versatile resource for research into German language processing, semantic memory, and cross-linguistic variation. By making the data set openly available, we aim to support both methodologically grounded stimulus selection and the development and evaluation of models of human semantic knowledge. Details about data access are provided in the section \textit{Availability of Data and Materials}.

\section{Acknowledgments}

We thank all the numerous volunteers who contributed to the project through participation in the free-association task and dissemination of the project. We thank Harald Baayen for contributions early in the project. Finally, we thank Laura Wiles for editing the manuscript.

\section{Declarations}

\subsection{Funding}

This work was supported by a grant from the Swiss National Science Foundation to Dirk U. Wulff (197315).
Kaidi Lõo acknowledges funding from the Estonian Research Council grant PSG743 (PI: Kaidi Lõo).

\subsection{Conflicts of Interest/Competing Interests}

The authors declare to have no conflicts of interest or competing interests.

\subsection{Ethics Approval}

This study was performed in line with the principles of the Declaration of Helsinki and was reviewed and approved by the institutional review board of the Faculty of Psychology at the University of Basel with the identifier 012-14-2. The project was further reviewed and approved by the institutional review board of the KU Leuven with the identifier G-2020-2140-R3(AMD).

\subsection{Consent to Participate}

Consent to participate was obtained by showing participants a page with information about the goals of the project, contents and duration of the web-based free-association task, and responsible researchers including links to contact them, followed by a button labeled ``I want to participate'', which all participants clicked prior to participation. Additionally, participants were informed that any data they provided would be stored anonymously and used for research purposes only.

\subsection{Availability of Data and Materials}

The data set generated and analyzed in this study is available for non-commercial purposes from the Small World of Words research website \href{https://smallworldofwords.org/de/project/research}{https://smallworldofwords.org/de/project/research}. The ``SWOW-DE'' data set analyzed here can be downloaded as ``SWOW-DE 2025 R55'' and includes three responses from exactly 55 participants for each of 5,877 cue words. In addition, a data set ``SWOW-DE 2025 RAW'' including all collected responses from included participants for 6,003 cue words (including some with few responses) is available. We recommend using the data found in ``SWOW-DE 2025 R55'' for most applications and would like to encourage the explicit mentioning of the exact data set used as well as citing this article when using it.

\subsection{Code Availability}

The code to perform preprocessing of the raw data and all analyses performed for this paper are available in the GitHub repository \href{https://github.com/samuelae/SWOW-DE-2025-Code}{https://github.com/samuelae/SWOW-DE-2025-Code}.

\subsection{Authors' Contributions}

Conceptualization: S.A., S.D.D., K.L. R.M., D.U.W.; 
Data curation: S.A., S.D.D., D.U.W.; 
Formal analysis: S.A.; 
Funding acquisition: R.M., S.D.D., D.U.W.; 
Investigation: S.A., R.M., D.U.W.; 
Methodology: S.A., K.L., S.D.D., D.U.W.; 
Project administration: S.A., R.M., S.D.D., D.U.W.; 
Resources: S.A., K.L., S.D.D., D.U.W.; 
Software: S.A., S.D.D., D.U.W; 
Supervision: R.M., D.U.W.; 
Visualization: S.A., D.U.W.; 
Writing – original draft: S.A.; 
Writing – review \& editing: S.A., R.M., K.L., S.D.D., D.U.W.

\printbibliography

\clearpage

\appendix

\section{Free-Association Task Instructions}
\label{app:instruction}

\begin{figure}[h]
    \includegraphics[width=\linewidth]{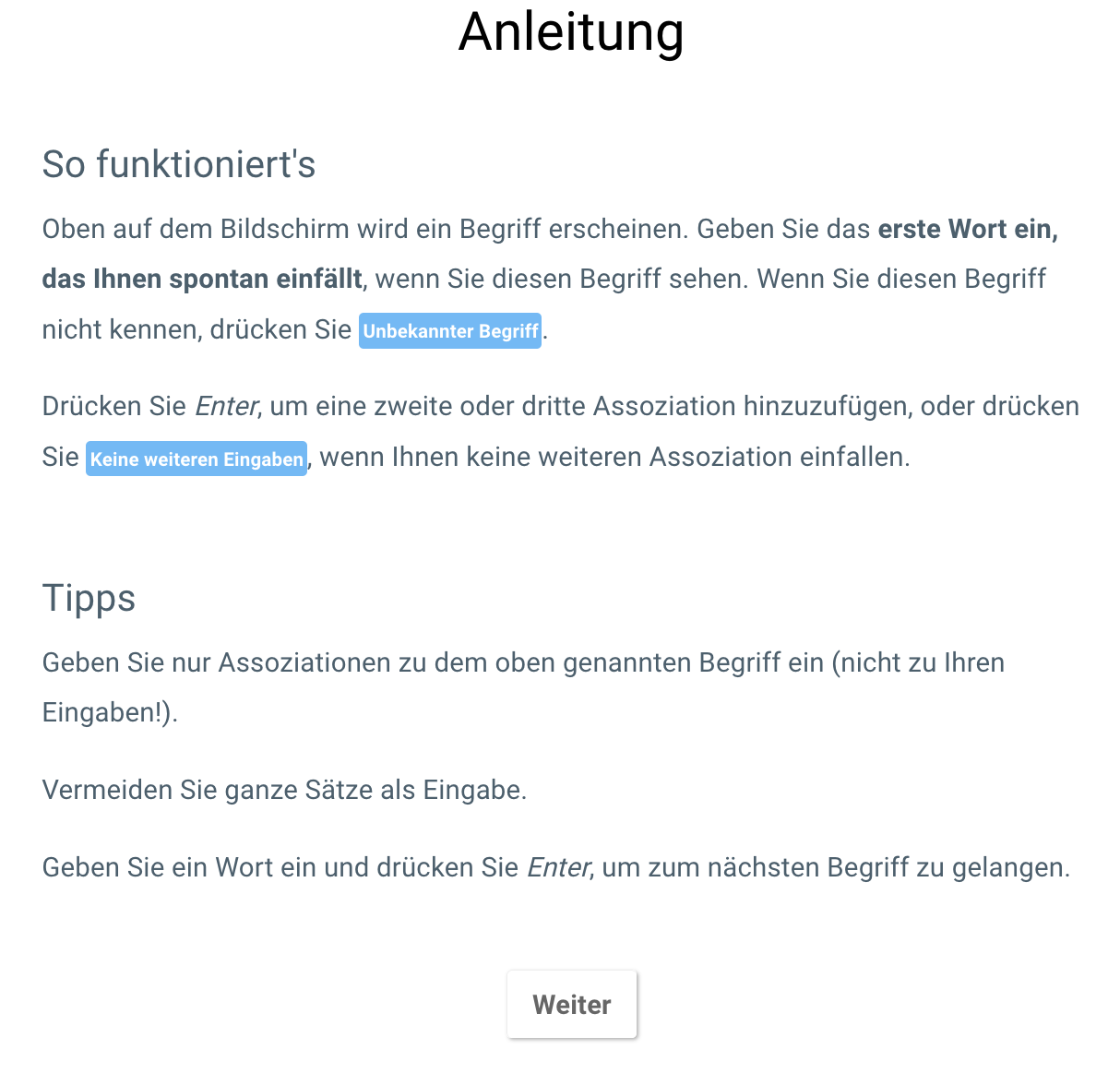}
    \caption{Free-association instructions. Original instructions in German, see English translation.}
    \label{fig:instruction}
\end{figure}

Figure~\ref{fig:instruction} shows original German instructions for the free-association task. English translation of the instructions for the free-association task:

\begin{center}
\textbf{Instruction}
\end{center}

\begin{flushleft}
\textbf{This is how it works}

On the top of the screen, a cue word will appear. Enter the first word that spontaneously comes to mind, when seeing this cue word. If you do not know the cue word, click on ``unknown word''. 

Press \textit{Enter}, to add a second and third association, or click ``no further entries'', if no further associations come to mind. 
\end{flushleft}

\begin{flushleft}
\textbf{Tipps}

Only enter associations to the cue word at the top of the screen (not to your own entries).

Avoid complete sentences as entries.

Enter a word and press \textit{Enter} to advance to the next entry.

\end{flushleft}

\section{Participant selection}
\label{app:participant_sel}

\begin{figure}[h]
    \includegraphics[width=0.48\textwidth]{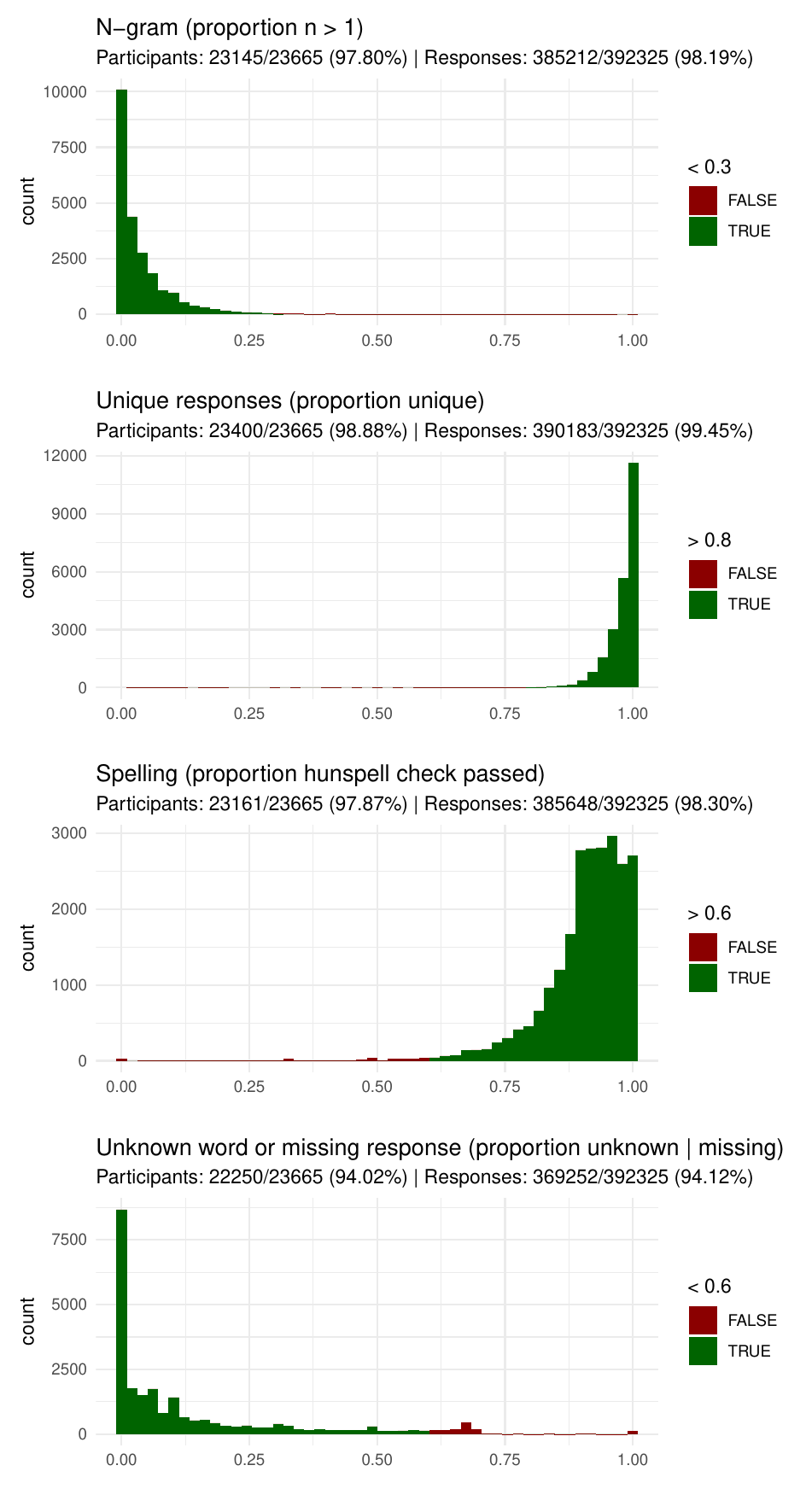}
    \caption{\textbf{Participant inclusion criteria.} The effect of individual participant inclusion criteria on participant selection.}
    \label{fig:part-sel}
\end{figure}

Figure~\ref{fig:part-sel} details each participant inclusion criterion's effect on the number of chosen participants. Note that some of the participants will fall short of multiple criteria.

\newpage

\section{LLM Spelling Improvement Prompts}
\label{app:prompts}

\subsection{Model and Settings}

We implemented the LLM-based spelling improvements using the models and settings in Table~\ref{tab:llms}. Please note that while both models were used in model evaluation, only ``gpt-oss:120b'' was used to improve the spelling in the SWOW-DE data.

\begin{table}[H]
  \begin{threeparttable}
      \caption{LLM Settings and Evaluation}
    \label{tab:llms}
\begin{tabular}{lccc}
\hline
Model & Temperature & Accuracy \\ 
\hline
llama4:16x17b "Scout" & 0 & .45 \\
gpt-oss:120b & 0 & .66 \\
\hline
\end{tabular}
\vspace{0.5em}
    \begin{tablenotes}[para,flushleft]
        {\small
            \textit{Note.} Except for temperature, both models were used with standard parameters. Find implementations at https://ollama.com/library/llama4 and https://ollama.com/library/gpt-oss.
         }
    \end{tablenotes}
\end{threeparttable}
\end{table}

\subsection{Prompts}

\noindent Prompts were developed in an iterative process using a random sample of 1,000 trials containing at least one misspelled response for which human corrections were obtained. Half of the sample was used to test different prompts and LLMs, while the other half served as an evaluation set to estimate the chosen setup's agreement with the human corrections independently of the sample originally used to develop the prompts. The prompts were all written in German, to keep the language of instructions in line with the contents.

\subsubsection{System Prompt}

\noindent The following system prompt was used for all spelling improvement LLM queries:

\begin{prompt}
Du hilfst mir potenziell falsch geschriebene Responses in einer Studie mit freien Assoziationen zu korrigieren. Ändere so wenig wie möglich und folge den Regeln. 

WICHTIG: Antworte ausschließlich mit der korrigierten Response, ohne Anführungszeichen oder Zusatztext. 

REGELN: 

1. Wenn eine Response a) richtig geschrieben ist auf Deutsch, b) richtig geschrieben ist in einer anderen Sprache und kein typisches, deutsches falsch geschriebenes Wort ist, oder c) ein korrekt geschriebener Eigenname ist: Die originale Response verwenden. 

Beispiele:
Cue: Mensch; Response: Homo sapiens; Korrigiert: Homo sapiens
Cue: Theorie; Response: The Big Bang Theory; Korrigiert: The Big Bang Theory

2. Wenn eine Response falsch geschrieben ist und (in Betracht des Cues und der anderen Assoziationen) die richtige Schreibweise zugerordnet werden kann: Die korrekt geschriebene Response verwenden.

Beispiele: 
Cue: Argument; Response: Diskussoin; Korrigiert: Diskussion
Cue: Marketing; Response: Werbungh; Korrigiert: Werbung

3. Wenn eine Response Wortkonstrukte enthält die auf mehrere Responses in einem Antwortfeld hinweisen: Die erste Response verwenden.

Beispiele:
Cue: hören; Response: sehen, fühlen, riechen; Korrigiert: sehen
Cue: Schleifpapier; Response: feinkörnig/grobkörnig; Korrigiert: feinkörnig

4. Wenn eine Assoziation Wortkonstrukte enthält, die den Cue wiederholen und eine zusätzliche, eigenständige Komponente enthalten: Die korrekt geschriebene eigenständige Komponente verwenden.

Beispiele:
Cue: lokal; Response: Lokal/Bar; Korrigiert: Bar

5. Wenn eine Response ein unvollständiges Wort ist aber in Kombination mit dem Cue Sinn macht: Die sinnvolle Kombination verwenden.

Beispiele:
Cue: Gebiet; Response: Hohheits; Korrigiert: Hoheitsgebiet
Cue: notwendig; Response: keit; Korrigiert: Notwendigkeit
\end{prompt}

\noindent English translation of the system prompt:

\begin{prompt}
You help me correct potentially misspelled responses in a free association study. Change as little as possible and follow the rules.

IMPORTANT: Reply exclusively with the corrected response, without quotation marks or additional text.

RULES:

1. If a response a) is correctly spelled in German, b) is correctly spelled in another language and is not a typical, misspelled German word, or c) is a correctly spelled proper name: Use the original response.

Examples:
Cue: Mensch; Response: Homo sapiens; Corrected: Homo sapiens
Cue: Theorie; Response: The Big Bang Theory; Corrected: The Big Bang Theory

2. If a response is misspelled and (considering the cue and the other associations) the correct spelling can be assigned: Use the correctly spelled response.

Examples:
Cue: Argument; Response: Diskussoin; Corrected: Diskussion
Cue: Marketing; Response: Werbungh; Corrected: Werbung

3. If a response contains word constructs that indicate multiple responses in one answer field: Use the first response.

Examples:
Cue: hören; Response: sehen, fühlen, riechen; Corrected: sehen
Cue: Schleifpapier; Response: feinkörnig/grobkörnig; Corrected: feinkörnig

4. If an association contains word constructs that repeat the cue and contain an additional, independent component: Use the correctly spelled independent component.

Examples:
Cue: lokal; Response: Lokal/Bar; Corrected: Bar

5. If a response is an incomplete word but makes sense in combination with the cue: Use the meaningful combination.

Examples:
Cue: Gebiet; Response: Hohheits; Corrected: Hoheitsgebiet
Cue: notwendig; Response: keit; Corrected: Notwendigkeit
\end{prompt}

\subsubsection{User Prompt}

\noindent For each incorrectly spelled response, the following prompt template was used to obtain a correction. 

\begin{prompt}
Responses zum Cue '{cue}' von dieser Person sind: 1. '{resp_1}', 2. '{resp_2}', 3. '{resp_3}'.
Die Person hat statt der Maske 'WORD' ursprünglich die Response '{incorrect}' hingeschrieben.
Wie sollte die Response '{incorrect}' korrigiert lauten?
Antworte nur mit der korrigierten Response!
\end{prompt}

\noindent English translation of the user prompt:

\begin{prompt}
Responses to the cue '{cue}' from this person are: 1. '{resp_1}', 2. '{resp_2}', 3. '{resp_3}'.
The person, instead of the mask 'WORD', originally wrote the response '{incorrect}'.
How should the response '{incorrect}' be corrected?
Respond exclusively with the corrected response!
\end{prompt}

\noindent For each LLM call, the user prompt was populated with the respective cue and responses of the trial where the prompt template contains variable names (e.g., \texttt{\{resp\_2\}}). For context the cue and all three responses are provided, but with the position containing the misspelled word masked using ``WORD''. Then the correction for the incorrectly spelled response (i.e., \texttt{\{incorrect\}}) is requested.

\section{SWOW-DE Word Embedding with German Labels}
\label{app:embedding-de}

\begin{figure*}[h]
    \includegraphics[width=\textwidth]{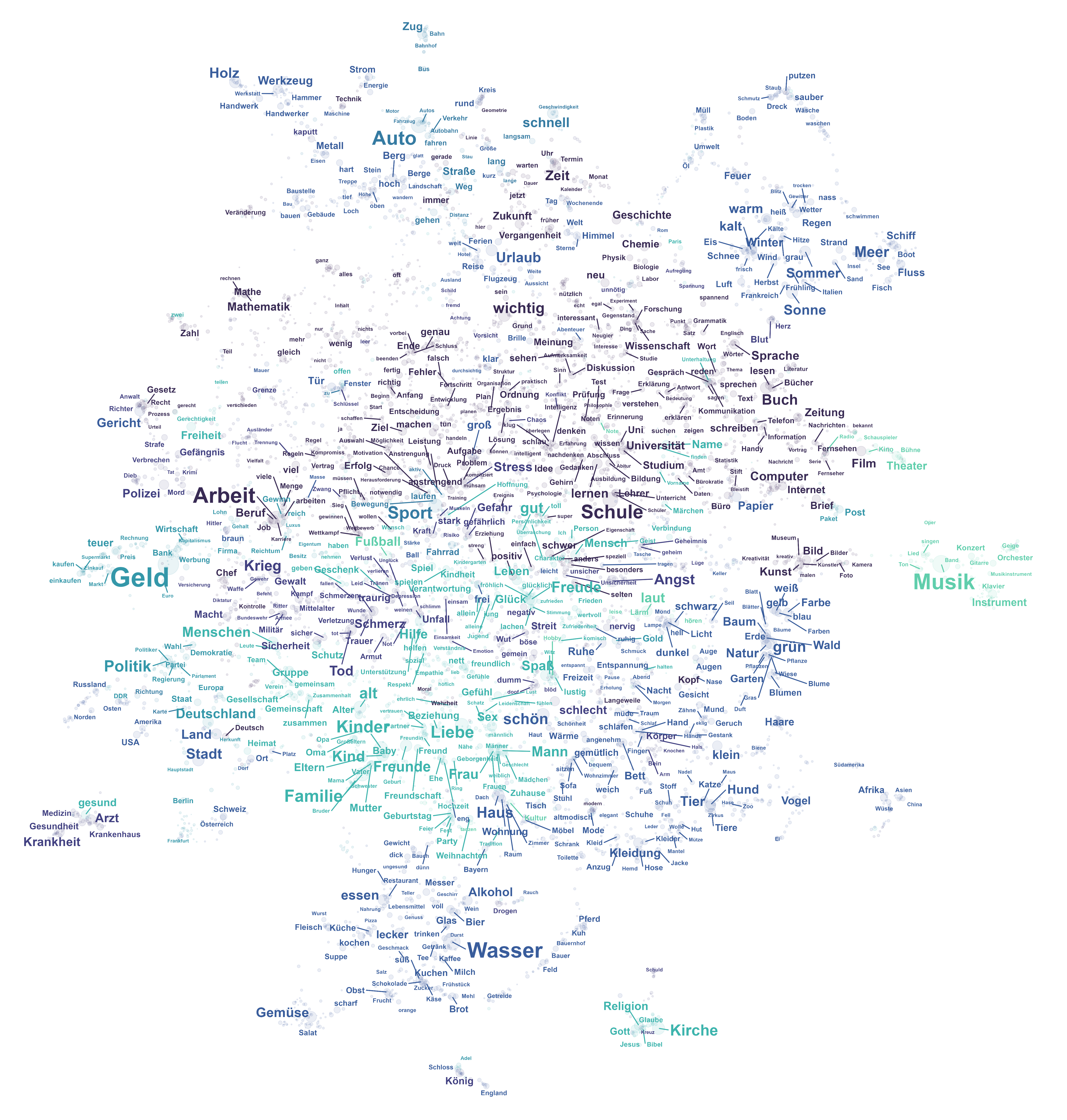}
    \caption{\textbf{Cue word embedding.} Projection of cue word embedding based on response frequencies using PPMI and SVD. Projection using UMAP \parencite{mcinnes_umap_2020} and colored by clusters detected with the Louvain clustering algorithm \parencite{blondel_fast_2008}. Larger points and labels indicate words more frequent among the SWOW-DE responses.}
    \label{fig:embedding-de}
\end{figure*}

\end{document}